\documentclass[twoside,11pt]{article}

\usepackage{jmlr2e}

\usepackage{caption}
\usepackage{subcaption}

\usepackage{todonotes}
\setuptodonotes{inline}
\usepackage{algorithm}
\usepackage{algpseudocode}
\usepackage{amsmath}
\usepackage{amssymb}
\usepackage{graphicx}
\usepackage{wrapfig}
\usepackage{placeins}
\usepackage{tcolorbox}
\usepackage{booktabs}
\usepackage{url}

\usepackage[capitalize]{cleveref}

\newcommand{\notw}{NO\textsubscript{2}}
\newcommand{\truepleb}{TruePLeBO}
\newcommand{\pleb}{PLeBO}
\newcommand{\dirtra}{DirectTrans}
\newcommand{\ransrch}{RandomSearch}
\newcommand{\shared}{Shared}
\newcommand{\D}{\boldsymbol{\mathcal{D}}}

\DeclareMathOperator*{\argmax}{arg\,max}
\newcommand{\xopt}{\xx^*}

\jmlrheading{1}{2023}{}{}{}{}

\ShortHeadings{Data-driven Prior Learning for Bayesian Optimisation}{Hellan, Lucas and Goddard}
\firstpageno{1}

\begin{document}

\title{Data-driven Prior Learning for Bayesian Optimisation}

\author{\name Sigrid Passano Hellan \email s.p.hellan@ed.ac.uk
        \AND 
        \name Christopher G. Lucas \email c.lucas@ed.ac.uk
        \AND
        \name Nigel H. Goddard \email nigel.goddard@ed.ac.uk \\ \\
       \addr School of Informatics\\
       University of Edinburgh, UK
}

\maketitle

\begin{abstract}
Transfer learning for Bayesian optimisation has generally assumed a strong similarity between optimisation tasks, with at least a subset having similar optimal inputs. This assumption can reduce computational costs, but it is violated in a wide range of optimisation problems where transfer learning may nonetheless be useful. We replace this assumption with a weaker one only requiring the shape of the optimisation landscape to be similar, and analyse the recent method \emph{Prior Learning for Bayesian Optimisation} --- \pleb{} --- in this setting. By learning priors for the hyperparameters of the Gaussian process surrogate model we 
can better approximate the underlying function, especially for few function evaluations.
We validate the learned priors and compare to a breadth of transfer learning approaches, using synthetic data and a recent air pollution optimisation problem as benchmarks.
We show that \pleb{} and prior transfer find good inputs in fewer evaluations.

\end{abstract}


\section{Introduction}

Our goal in Bayesian optimisation (BO) is to find high-performing input values using few function evaluations. A powerful approach is to transfer knowledge from previous optimisation tasks.
For example, when optimising many similar problems we can choose to disregard input ranges that consistently yield suboptimal outputs. Within the BO framework there are many ways of transferring knowledge \citep{bai2023transfer}, for instance through a shared model \citep{swersky2013multi}, through the initial input values which we evaluate \citep{feurer2015initializing}, or through limiting the search space \citep{perrone2019learning}. 
In this paper we develop the idea of prior transfer further, and introduce an extension to a recent prior transfer method \citep{hellan_bayesian_2022}, which we call Prior Learning for BO (\pleb{}).

\textbf{Direct transfer:} Most transfer learning for Bayesian optimisation rests on the assumption that inputs that are optimal for (some of) the tuning tasks will also be near-optimal for the new test task. For instance, a common technique is to evaluate a few good input values from previous tasks on a new task before doing Bayesian optimisation.

\textbf{Prior transfer:} An alternative to direct transfer is to assume only that the optimisation tasks have similar response surfaces, i.e. the shapes of their optimisation landscapes. A simple example is to assume that all optimisation tasks can be modelled using a single shared set of Gaussian process (GP) hyperparameter (HP) values, as is done in \cite{wang2022pre}. Another approach is to learn a feature transformation which is applied to the input to the GP before the covariance kernel. This is done in Few-Shot BO, an example of deep kernel learning where a neural network is used for this feature transformation \citep{wistuba2021few}. 
Hand-crafted HP priors can also be seen as prior transfer: with \pleb{} we automate this process.
\cite{muller2023pfns4bo} also learn HP priors from related data, but pick the best from a randomly sampled set instead of learning a posterior distribution.

\begin{figure}[h]
    \centering
    \includegraphics[width=0.95\linewidth]{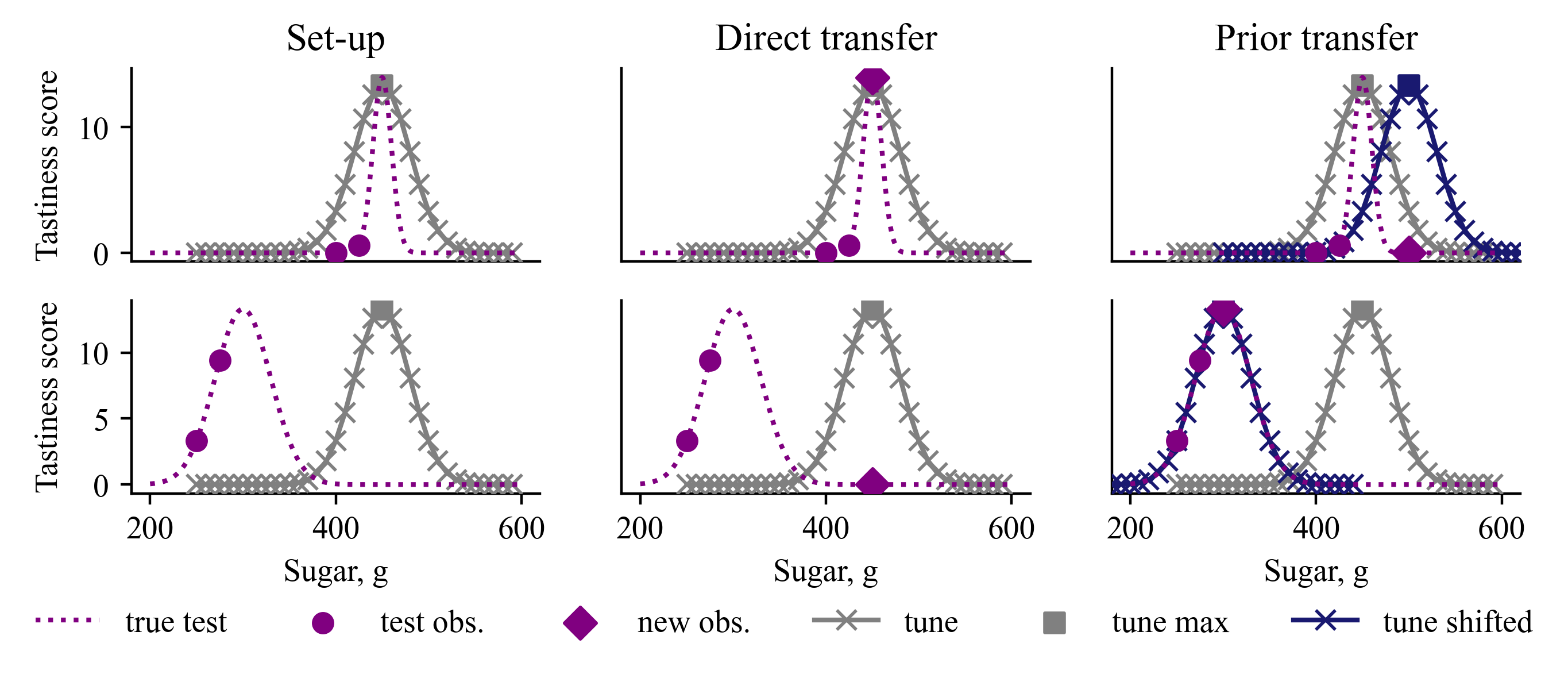}
\caption{Comparison of direct and prior transfer. 
In the top row the past and new tasks share optima but have different shapes. In the bottom row they instead have the same shape. 
Direct transfer works well for shared optima, prior transfer for shared shapes. 
}
\label{fig:compapre-prior-direct}
\end{figure}

\cref{fig:compapre-prior-direct} illustrates the difference between direct and prior transfer in two idealised settings. In the top row the optima for the tuning and test tasks coincide, and only direct transfer is able to perform useful transfer. In the bottom row the optima are far apart but the shapes are the same. In this setting prior transfer works perfectly while direct transfer does not work at all. For direct transfer we use a strategy where we evaluate the best point from the tuning data. 
We illustrate prior transfer with a
method where we assume the shapes are identical --- in practice we just assume the GP hyperparameters are similar. 

Most work on Bayesian optimisation and transfer learning has been applied to hyperparameter optimisation for machine learning (HPO), e.g. \cite{feurer2015initializing,perrone2019learning,wistuba2021few}. 
Extending the range of applications motivates the development of new methods, as assumptions that make sense for one application 
do not necessarily make sense for a different application. 
For HPO we can expect similar learning rates to work well for the same model trained on different data sets.
But in the air pollution context we do not expect the same coordinates in different cities to be the most polluted --- e.g. three kilometres north of the city centre. Therefore, we expect prior transfer to work well in situations unsuited to direct transfer, such as air pollution monitoring.

An earlier version of \pleb{} was presented as part of \cite{hellan_bayesian_2022}. Here we improve on the method and compare it to other transfer learning approaches.
We also provide code for a new and simpler implementation, and analyse the quality of the resulting priors. We show the advantage of prior transfer over direct transfer on both synthetic and real-world data, the latter being air pollution measurements.

\section{Bayesian optimisation}

\newcommand{\btau}{\boldsymbol{\tau}}
\newcommand{\bbea}{\boldsymbol{\eta}}
\newcommand{\bbta}{\boldsymbol{\theta}}

Bayesian optimisation \citep{garnett2023bayesian} is a blackbox optimisation technique, which utilises a surrogate model, typically a Gaussian process, to model the underlying optimisation task and come up with informed choices of input values. It is particularly suited to problems with a complex unknown structure, where we do not know the gradients, and which are expensive to evaluate. 
Gaussian processes \citep{rasmussen_gaussian_2006} are probabilistic machine learning models which are popular in the regime of small data. They are defined by their choice of mean and covariance functions. The former is often set to zero \citep{ath2020mean}, while the latter comes with hyperparameters that need to be set. For instance, the RBF kernel 
${k_{RBF}(\btau) = \sigma_r^2 \exp \left(- \frac{\btau^T\btau}{2l^2} \right)}$
has two hyperparameters: the lengthscale $l$ and the signal variance $\sigma_r^2$. The former defines how quickly the function changes, the latter how large its absolute values are. 
We also need an acquisition function, which uses the surrogate model to determine the next sampling location.
We use expected improvement (EI, \citet{jones_efficient_1998}) unless otherwise noted.

\newcommand{\xx}{\boldsymbol{x}}

The problems we want to optimise in Bayesian optimisation are of the form $\xopt = \argmax_{\xx} \: f({\xx})$. To fit the hyperparameters of the Gaussian process we use the log marginal likelihood. A simple approach is to use gradient-based optimisation, possibly including a prior term on the hyperparameters. Alternatively, we can approximate the posterior distribution of the hyperparameters, for instance using Markov chain Monte Carlo (MCMC) \citep{lalchand_approximate_2020}. Then we take the uncertainty of our hyperparameters into account in our model.
With transfer learning we assume access to historical evaluations of other tasks, changing the problem to $\xx^*_{j} = \argmax_{\xx} f_{j}(\xx) \:\: | \:\: \{\{\xx_{n,i},y_{n,i}\}_{i=1}^{N_n}\}_{n=1}^{N} $
where $n$ indexes tasks, $i$ indexes samples and $N_n$ is the number of sample evaluations for task $n$.

\section{\pleb{} } \label{plebo-method}

\begin{wrapfigure}{r}{8.5cm}
    \centering
    \includegraphics[width=0.98\linewidth]{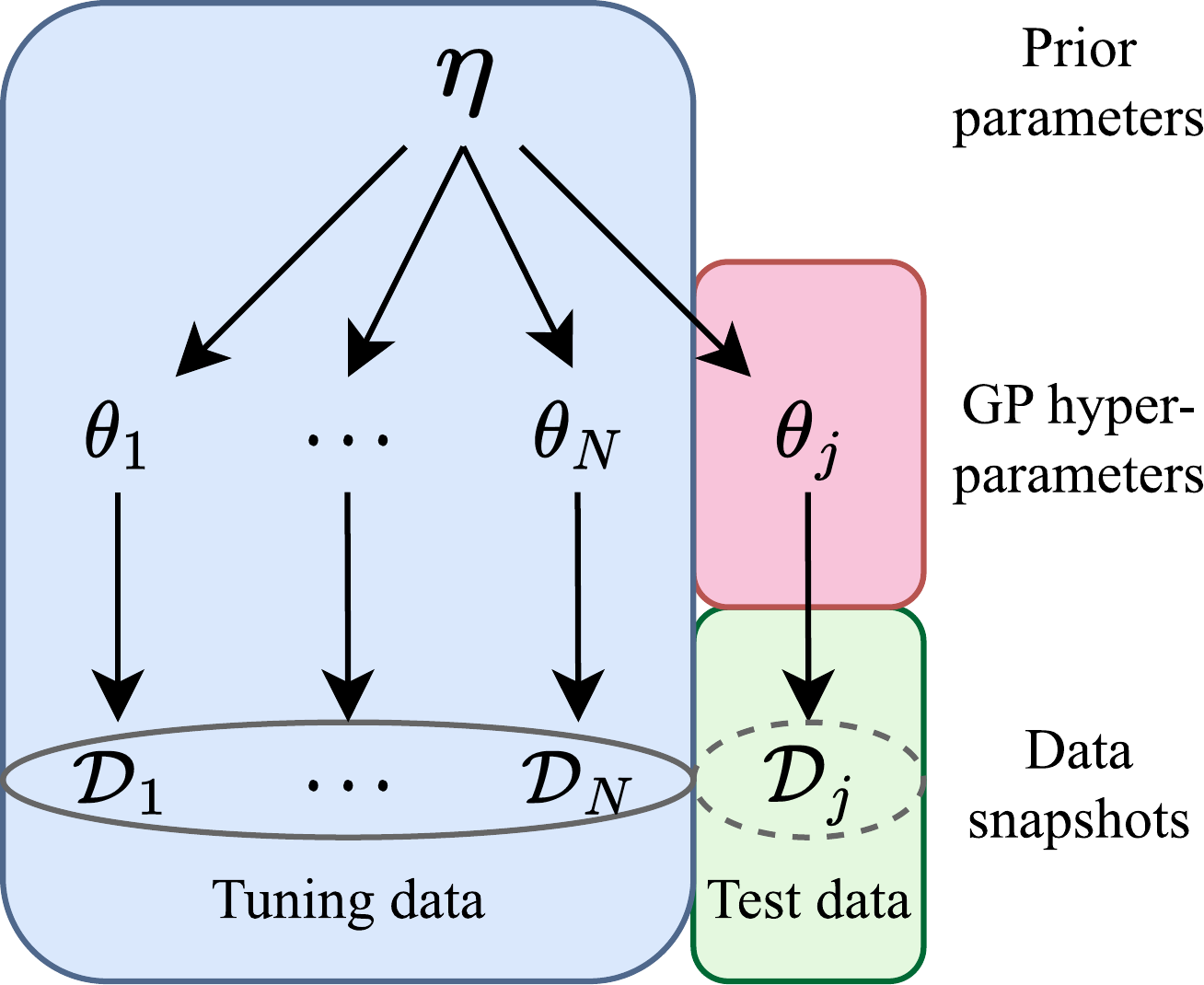}
    \caption{Hierarchical structure and inference in \pleb{}.}
    \label{fig:hierarchical-model}
\end{wrapfigure}

\definecolor{myLBlue}{RGB}{218, 232, 252}
\definecolor{myLRed}{RGB}{248, 196, 218}
\definecolor{myLGreen}{RGB}{228, 248, 223}

The \pleb{} method is an example of prior transfer, intended to work even with a small amount of tuning data. We assume a hierarchical structure where each optimisation task has its own set of GP hyperparameters $\bbta$, and these hyperparameters come from a shared distribution $\bbea$. The method has two parts: in the preprocessing step, we use \colorbox{myLBlue}{MCMC} to learn $\bbta$ and $\bbea$ given observations $\D$, and then \colorbox{myLRed}{sample new values} of $\bbta$ based on $\bbea$ to generate a set of candidate hyperparameters. In the \colorbox{myLGreen}{optimisation phase} we use importance weighting to fit these candidates to the observed test data $\D_j$. This structure is illustrated in \cref{fig:hierarchical-model}.

Formally, the structure is
$ p(\bbea,\bbta,\D)\!~=~\!p(\bbea)p(\bbta|\bbea)p(\D|\bbta)\!~=~\!p(\bbea)\prod_np(\bbta_n|\bbea)p(\D_n|\bbta_n) $. 
The modelling distributions can be chosen separately for each hyperparameter, we use gamma distributions $p(\bbta|\bbea)=\Gamma(\bbta;\bbea)$. That means we need to learn $N$ sets of $\bbta_n$ and 2$k$ parameters of $\bbea$, where $k$ is the dimensionality of $\bbta_n$ and $N$ is the number of tuning tasks. For simplicity, we restrict ourselves to 2 HPs in $\bbta_n$, so learn 4 parameters of $\bbea$. We also need to choose the number $H$ of candidate samples; we use 200 to get a good approximation to taking the expectation given $\bbea$, but future work should explore reducing $H$.

\textbf{Preprocessing:} We  learn distributions for the surrogate model hyperparameters, using Markov chain Monte Carlo to collect samples from the distributions of $\bbea$ and $\bbta$. 
We replace \citeauthor{hellan_bayesian_2022}'s (\citeyear{hellan_bayesian_2022}) custom-built implementation with a new one (\url{https://github.com/sighellan/plebo}) based on NumPyro \citep{phan2019composable} and using the NUTS \citep{hoffman_no-u-turn_2014} sampler.
It is more stable and makes modelling changes, e.g. changing $p(\bbta|\bbea)$, much simpler. 
As the sampling chains do not always find samples with non-zero likelihood we run multiple chains in parallel before a simple filtering postprocessing step.

\newcommand{\Dji}{\D_j^{:i}}
\newcommand{\Eop}{\mathop{\mathbb{E}}}
\newcommand{\tacand}{\bbta^\mathrm{cand}}
\newcommand{\acqbase}{a_i^\theta}

\textbf{Optimising:} We use the candidate samples within the optimisation loop to calculate the acquisition function, $a_{i}(\boldsymbol{x}) \approx \frac{1}{W} \sum_{h=1}^H  w_h \: \acqbase(\boldsymbol{x}, \tacand_h)$
where $w_h=p(\Dji|\tacand_h)$, $W~=~\sum_{h=1}^H~w_h$ and $\acqbase(\cdot)$ is the base acquisition function, for which we use EI. At each optimisation step we calculate the likelihood of every hyperparameter candidate, and use it to weight the corresponding base acquisition function. More details are in \cref{app:optimisation}.

\section{Experiments}

\newcommand{\taskmax}{y^{(j)}_\mathrm{max}}
\newcommand{\basemet}{r^{(j)}_i}
\newcommand{\nstart}{N_\mathrm{start}}
\newcommand{\obsvals}{\boldsymbol{Y}^{(j)}_{:\nstart+i}}

We consider both a synthetic benchmark and real-world air pollution data \citep{copernicus_short}. To simplify analysis we use the RBF kernel and limit ourselves to two hyperparameters: the lengthscale and the signal variance. 
Our base metric is the best observed value at each iteration $i$, normalised by the max value of the task, $\taskmax$: 
$\basemet = \max \obsvals / \taskmax$
where $\obsvals$ are the observations and $\nstart$ is the number of starting evaluations.

We compare \pleb{} to a collection of baselines with no, direct and prior transfer:
\begin{itemize}
    \setlength\itemsep{-0.5em}
    \vspace{-0.2cm}
    \item No transfer
\begin{itemize}
\vspace{-0.2cm}
    \setlength\itemsep{-0.2em}
    \item \textbf{\ransrch{}:} Randomly select the inputs at each iteration. 
    \item \textbf{BoTorch:} Default settings from the BoTorch library \citep{balandat2020botorch}.
    \item \textbf{EI:} BO without transfer using expected improvement \citep{jones_efficient_1998}.
    \item \textbf{UCB:} BO without transfer using upper confidence bound  \citep{srinivas_gaussian_2012}.
\end{itemize} 
\item Direct transfer
\begin{itemize}
\vspace{-0.2cm}
    \setlength\itemsep{-0.2em}
    \item \textbf{\dirtra{}:} Shared GP for past and new tasks. 
    Cap at 100 past evaluations.
    \item \textbf{Initial:} First evaluate the best point from each previous task, then do \textbf{EI}.
\end{itemize}
\item Prior transfer
\begin{itemize}
\vspace{-0.2cm}
    \setlength\itemsep{-0.2em}
    \item \textbf{\pleb{}:} The new prior learning method detailed in \cref{plebo-method}.
    \item \textbf{\truepleb{}:} Using EI with the true hyperparameters (only for synthetic data).
    \item \textbf{Gamma:} Use the mean prior $\bbea$ from \pleb{} when optimising the HPs. 
    \item \textbf{\shared{}:} A single set of HPs learned from the tuning tasks
    \citep{wang2022pre}.
\end{itemize}
\end{itemize}

\begin{figure}[h]
\centering
\includegraphics[width=0.8\linewidth]{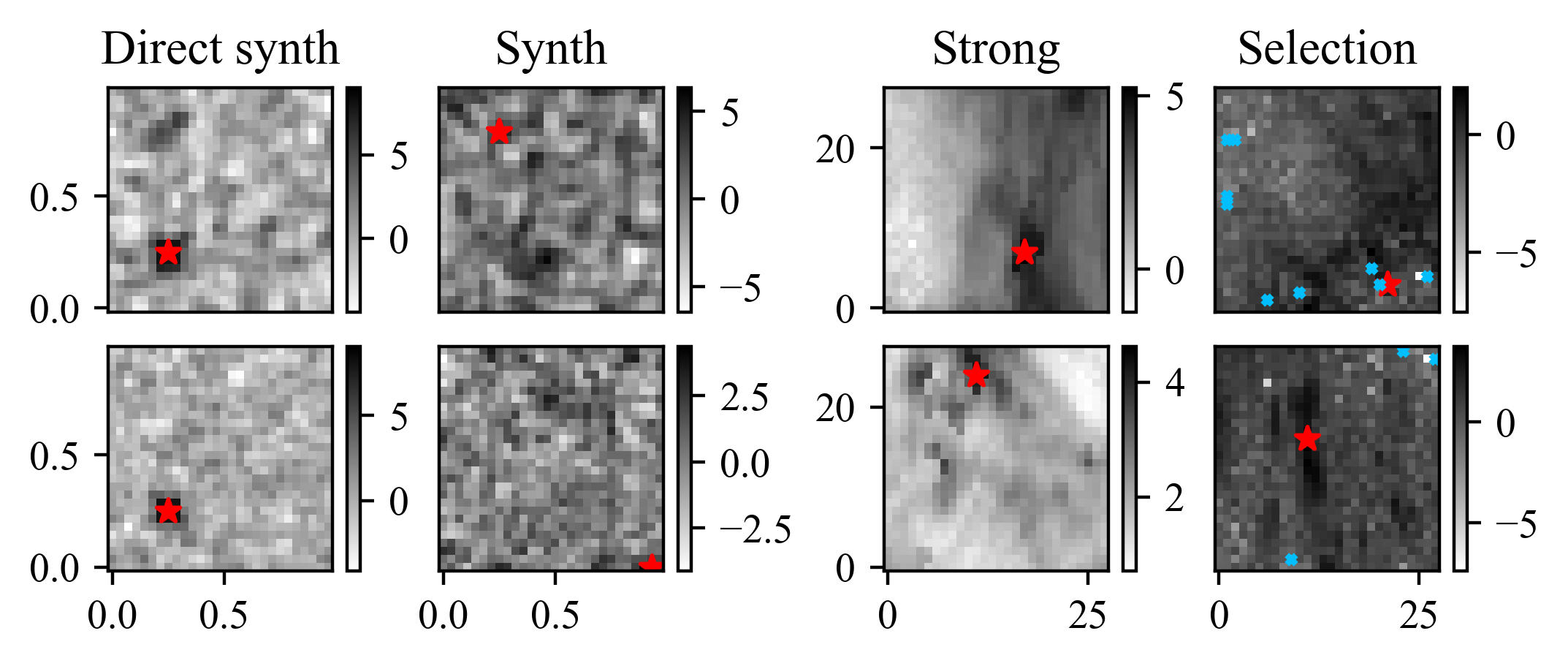}
\caption{Example optimisation tasks. The left column shows the standard assumption that optima are near each other. Stars indicate optima and blue crosses missing values.}
\label{fig-example-tasks}
\end{figure}

The \textbf{synthetic benchmark} allows us to evaluate the extracted priors. 
We use the same hierarchical model as in \pleb{}, see \cref{fig:hierarchical-model}. We use GPs with RBF kernels and a known low level of noise. We define gamma distributions on the two hyperparameters: the lengthscale $l \sim \Gamma(5, 0.01)$, and the signal variance $\sigma_r^2 \sim \Gamma(2, 2)$. We use the shape, scale parameterisation of $\Gamma(\cdot)$. 
We generate a set of 10 tuning optimisation tasks and 100 test tasks. The tuning tasks have 20 evaluations each, and we start the test tasks with ten known evaluations. The tasks are discretised to match the other benchmarks. 
See \cref{fig-example-tasks}.

\begin{figure}[h]
\centering

\includegraphics[width=0.98\linewidth]{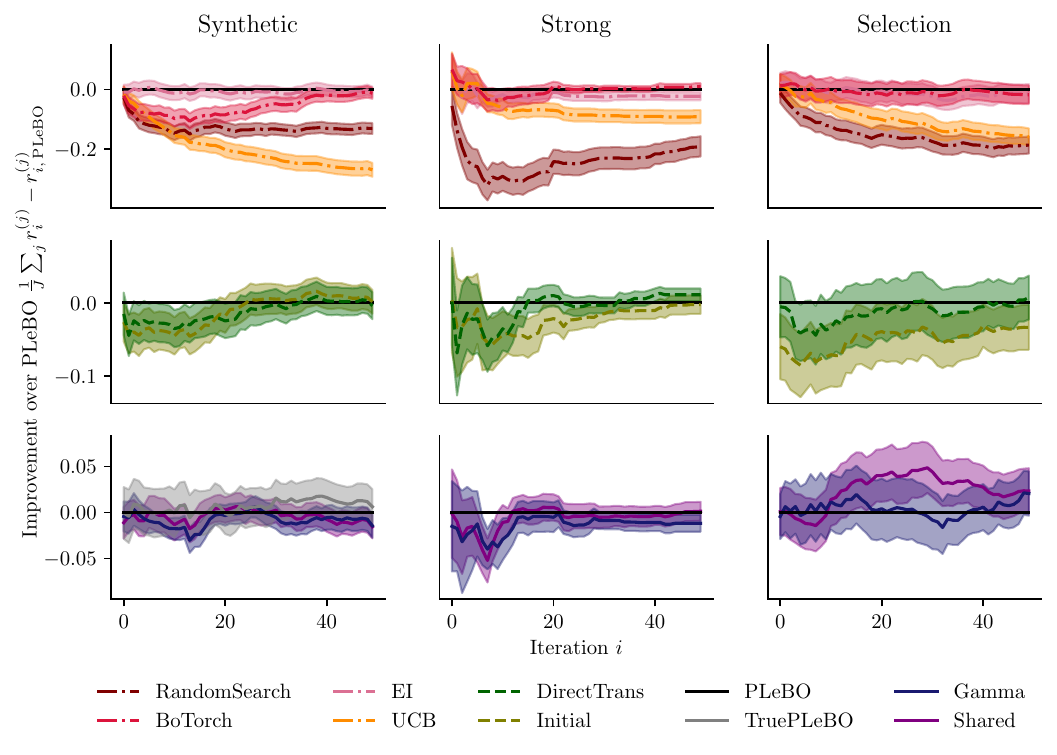}
\caption{Performance compared to \pleb{} at same iteration (above zero is improvement), mean $\pm$ one standard error. $J$ is the number of test tasks.
The top row compares \pleb{} to no transfer, the middle to direct transfer and the bottom row to prior transfer methods. 
}
\label{fig-plebo-results}
\end{figure}

We also use two \textbf{air pollution benchmarks}: the satellite optimisation tasks used in \cite{hellan_optimising_2020_arxiv}. 
Each task comes from a snapshot of \notw{} air pollution levels taken from the Sentinel-5P satellite of the Copernicus programme \citep{copernicus_short}.
The two benchmarks come from grouping the snapshots based on the maximum pollution present. The Strong benchmark consists of 50 test tasks and 10 tuning tasks of similar problems. The Selection benchmark consists of 100 test tasks and 10 tuning tasks with maximal pollution levels at the middle of those of the test tasks. 
It therefore checks whether prior transfer still works if the task distributions change between the tuning and test data. 
The data has been preprocessed by log transform and standardisation, see \cite{hellan_optimising_2020_arxiv}.

\section{Results and Discussion}

\cref{fig-plebo-results} compares the performance of \pleb{} to the other methods. We see that the transfer methods generally find good input values in fewer iterations. EI only does slightly worse than \pleb{}. This highlights the difficulty of prior transfer, as we still need to explore the search space.
We also see that prior transfer requires fewer iterations than direct transfer.
The differences are smaller among the prior transfer methods. \pleb{} is better than \shared{} and Gamma on the synthetic and Strong benchmarks, but for the Selection benchmark \shared{} works best. 
That \truepleb{} works best on the synthetic data supports our  intuition that the optimisation can be sped up by better surrogate model HP estimates.

To validate \pleb{} we also analyse the inferred priors. In \cref{app-prior-quality} we compare the inferred and true values for $\bbea$ and $\bbta$. While the true and estimated priors are not identical they are similar, and an exact reconstruction could not be expected from only 10 tuning tasks. We also analysed the fits on the individual tasks: the true and inferred values had very similar likelihoods, and sometimes the inferred values had higher likelihood.

\pleb{} is computationally more expensive than the baselines. Its preprocessing step has a variable runtime due to the NUTS sampler. The runtime was 9 minutes for the synthetic, and 2.6--3.4 hours for the air pollution benchmarks on a standard desktop. The preprocessing is only done once, so a higher computational expense can be tolerated.
In the optimisation step we replace HP optimisation with calculating the acquisition function for each of $H$ candidate HP sets. This scales as O($H(i+\nstart)^3$). On average, \pleb{} takes about 6.7 seconds per optimisation step, compared to 0.13 seconds for EI. For very expensive optimisation tasks this is still only a fraction of the total cost, e.g. when installing an air pollution sensor. Runtimes for all methods are given in \cref{app-runtime}. 
\pleb{}'s runtime can be adjusted through $H$ based on the available budget and task evaluation speed.

We have presented an improved prior transfer method, \pleb{},
and evaluated it on a real-world benchmark for air pollution monitoring. 
We showed that the generated priors align with the tuning data, and can be exploited on new tasks. The approach is modular, and is not tied to a specific acquisition function. 
While more general, prior transfer is also weaker than direct transfer as we do not learn about the optima directly:
we showed some improvement over EI, but not a massive difference. 
Future work should attempt to exploit the correct HP choices more, and evaluate \pleb{} on a wider range of benchmarks.


\acks{
We thank Tiffany Vlaar and Linus Ericsson for comments on the paper, and G-Research for generous financial support towards conference attendance.

This work was supported by the EPSRC Centre for Doctoral Training in Data Science, funded by the UK Engineering and Physical Sciences Research Council (grant EP/L016427/1) and the University of Edinburgh.
}

\vskip 0.2in
\bibliography{zotero_local_local,background,boacc_unique_local}


\newpage

\appendix
\section{Supplementary material}
\label{app:suppl}

\subsection{Ethics statement}

The work uses no personal data and the goal of the application is to reduce the impact of air pollution on human health. The authors have no ethical concerns. 

\subsection{\pleb{} optimisation step} \label{app:optimisation}

\cref{alg:plebo} gives the procedure for calculating the acquisition function which is followed at each optimisation step. This is also expressed in \cref{eq-acq-plebo}, which shows how this corresponds to the expected acquisition function when drawing the hyperparameters from their posterior distribution. $\D_{1:N}$ is the set of tuning observations. $\Dji$ is the test observations available at iteration $i$. 

\newcommand{\bbaa}{\boldsymbol{a}}

\begin{algorithm}
\caption{\pleb{} acquisition calculation}\label{alg:plebo}
\begin{algorithmic}
\State Given $\tacand$, $\Dji$, $\boldsymbol{X}$, $\acqbase(\cdot)$
\State $\bbaa,W \gets \boldsymbol{0},0$
\For{$h$ in $1, \ldots, H$}
\State $w \gets p(\Dji|\tacand_h)$
\State $\bbaa \gets \bbaa + w \: \acqbase(\boldsymbol{X}, \tacand_h)$
\State $W \gets W + w$
\EndFor
\State Return $\bbaa/W$
\end{algorithmic}
\end{algorithm}

\begin{equation}
    a_{i}(\boldsymbol{x})~=~\Eop_{p(\tacand_h|\D_{1:N},\Dji)} [\acqbase(\boldsymbol{x}, \tacand_h)]\approx \frac{1}{W} \sum_{h=1}^H  w_h \: \acqbase(\boldsymbol{x}, \tacand_h), \:\: \tacand_h \sim p(\tacand_h|\bbea) \label{eq-acq-plebo}
\end{equation}

\subsection{Runtime} \label{app-runtime}

\cref{tab:app-pre-runtime} gives the runtime for the preprocessing step of \pleb{} for each benchmark. For the air pollution benchmarks we use 100 evaluations from each of the tuning tasks. \cref{tab:app-opt-runtime} gives the mean duration of each optimisation step for all the considered methods. We see that \pleb{} is much slower than the other methods. But for expensive optimisation tasks, where each step corresponds to e.g. installing a pollution sensor or training a neural network, this cost is negligible. We have capped the number of past evaluations available to \dirtra{} at 100, otherwise it would be much slower than the listed times.

The optimisation step of \pleb{} scales as O($H(i+\nstart)^3$), as opposed to \\ O($N_\mathrm{grad}(i+\nstart)^3)$ for EI where $N_\mathrm{grad}$ is the number of gradient steps \citep[p.~19]{rasmussen_gaussian_2006}. $H$ is a hyperparameter of the method which we can adjust to balance computational cost with the quality of the posterior estimate of the GP hyperparameters. At one extreme we could set $H$ to 1, which should give a similar runtime to \shared{}.

\begin{table}[h]
\centering
\begin{tabular}{lr}
\hline
Benchmark & Preprocessing time \\ \hline
Synthetic & 9 min 11 sec       \\
Strong    & 2 h 38 min         \\
Selection & 3 h 22 min         \\ \hline
\end{tabular}
\caption{\pleb{} preprocessing times.}
\label{tab:app-pre-runtime}
\end{table}

\begin{table}[h]
\centering
\begin{tabular}{@{}lrrr@{}}
\toprule
Method       & Synthetic & \multicolumn{1}{l}{Strong} & \multicolumn{1}{l}{Selection} \\ \midrule
\ransrch{} & 0.006      & 0.005                      & 0.005                         \\
BoTorch      & 0.20      & 0.31                       & 0.27                          \\
EI           & 0.16      & 0.11                       & 0.11                          \\
UCB          & 0.17      & 0.11                       & 0.11                          \\ \midrule
\dirtra{}  & 1.35      & 1.20                       & 1.20                          \\
Initial      & 0.16      & 0.10                       & 0.11                          \\ \midrule
\pleb{}        & 6.73      & 6.60                       & 6.89                          \\
\truepleb{}    & 0.05      & -                           & -                              \\
Gamma        & 0.18      & 0.13                       & 0.12                          \\
\shared{}       & 0.05      & 0.04                       & 0.04                          \\ \bottomrule
\end{tabular}
\caption{Mean durations in seconds of an optimisation step.}
\label{tab:app-opt-runtime}
\end{table}

\subsection{Prior quality} \label{app-prior-quality}

\begin{figure}[h]
     \centering
     \begin{subfigure}[b]{0.48\textwidth}
         \centering
         \includegraphics[width=\textwidth]{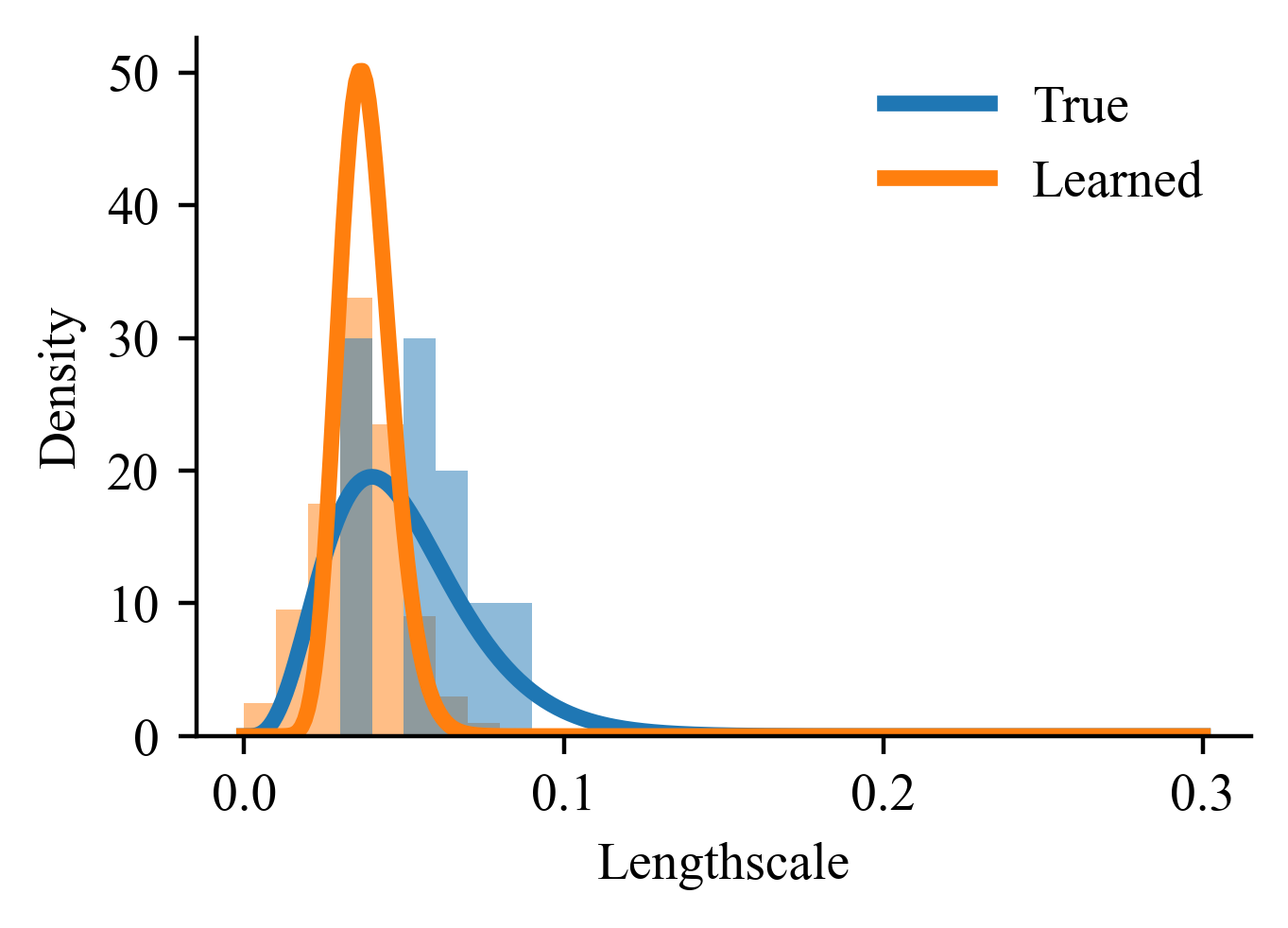}
     \end{subfigure}
     \hfill
     \begin{subfigure}[b]{0.48\textwidth}
         \centering
         \includegraphics[width=\textwidth]{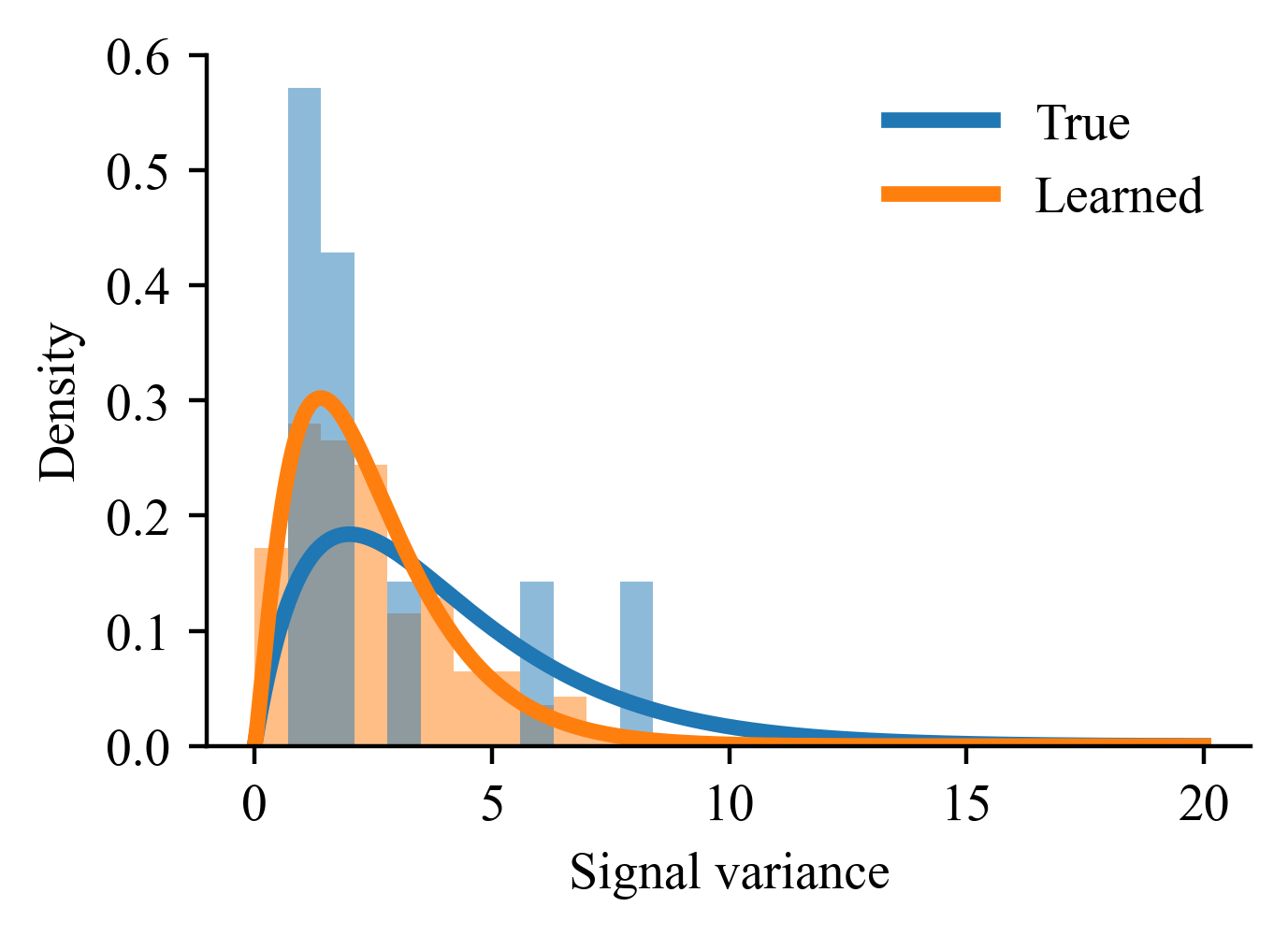}
     \end{subfigure}
        \caption{Comparing true and inferred hyperparameter priors for the synthetic benchmark.}
        \label{fig:plebo-synth-prior-fit}
\end{figure}

\cref{fig:plebo-synth-prior-fit} compares the true and inferred values of $\bbea$ and $\bbta$ for the lengthscale and signal variance. The learned distributions are more peaked than the true ones, but are similar and reasonable approximations given that we only use 10 tuning tasks.

\end{document}